%% file: main.tex
\documentclass[10pt,twocolumn,letterpaper]{article}

\usepackage[pagenumbers]{cvpr}%

\input{preamble}

\input{gen/nums}
\definecolor{cvprblue}{rgb}{0.21,0.49,0.74}
\usepackage[pagebackref,breaklinks,colorlinks,allcolors=cvprblue]{hyperref}

\def\confName{CVPR}
\def\confYear{2027}

\title{
ReCal-VLA: Sample-Efficient VLA Reinforcement Learning via Replay-Calibrated World-Model Mixing}

\makeatletter
\author{
Xiaowen Chang\textsuperscript{1,2}\thanks{Equal contribution.} \quad
Haoming Xu\textsuperscript{1,2}\footnotemark[1] \quad
Yiqi Deng\textsuperscript{2,3} \quad
Yaqing Xu\textsuperscript{2,4} \\
Xiurui Guo \textsuperscript{2,5} \quad
Wenchao Xue \textsuperscript{1}\thanks{Corresponding authors:~\begin{tabular}[t]{@{}l@{}}wenchaoxue@amss.ac.cn,\\xujiafeng26@stu.pku.edu.cn.\end{tabular}} \quad
Jiafeng Xu\textsuperscript{6}\footnotemark[2] \quad
Hao Dong\textsuperscript{2,6}\protected@xdef\@thanks{\unexpanded{\begingroup\gdef\@thefnmark{}\@footnotetext{Work done during internship at QYRO.}\endgroup}\unexpanded\expandafter{\@thanks}}\\[0.5em]
\textsuperscript{1}University of Chinese Academy of Sciences \quad
\textsuperscript{2}QYRO \quad
\textsuperscript{3}Beihang University \\
\textsuperscript{4}Renmin University of China \quad
\textsuperscript{5}Beijing Institute of Technology \quad
\textsuperscript{6}Peking University\\[-6em]
}
\makeatother

\begin{document}
\maketitle
\input{fig/teaser}
\input{sec/0_abstract}
\input{sec/1_intro}
\input{sec/2_related}

\input{sec/3_consistency}

\input{sec/4_method}
\input{sec/5_experiments}

\input{sec/5b_pro_real}

\input{sec/6_conclusion}
{
    \small
    \bibliographystyle{ieeenat_fullname}
    \bibliography{main}
}
\end{document}

%% file: preamble.tex
\usepackage{cuted}%

\usepackage{currfile}%

\usepackage{caption}%

\usepackage{multirow}
\usepackage{xspace}
\newcommand{\ours}{\textsc{ReCal-VLA}\xspace}

\usepackage{tikz}
\usetikzlibrary{arrows.meta,positioning,calc}

%% file: fig/teaser.tex
\begin{strip}
\centering
\includegraphics[width=0.96\linewidth]{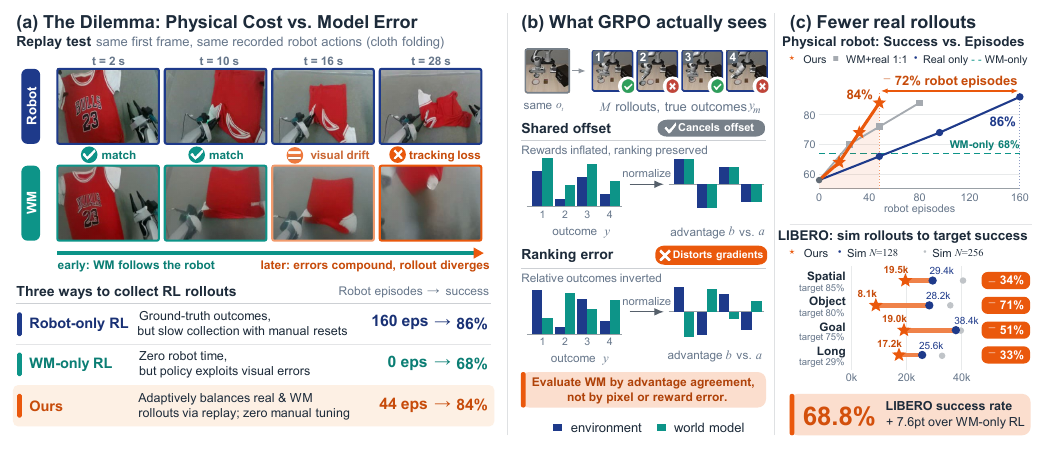}
\captionsetup{hypcap=false}
\captionof{figure}{\textbf{Where RL rollouts come from, and what a world model must get right.}
(a)~Replaying recorded robot actions from the same first frame, the WM follows the robot early on and then drifts; robot-only RL is grounded but costly, WM-only RL is cheap but exploitable, and \ours mixes both through replay calibration.
(b)~Under group-relative normalization, a reward offset shared by a group cancels, whereas a within-group error in relative outcomes flips advantage signs.
(c)~Robot episodes (top) and simulator trajectories (bottom) needed to reach a given success rate.}
\label{fig:teaser}
\captionsetup{hypcap=true}
\end{strip}

%% file: sec/0_abstract.tex
\begin{abstract}

Reinforcement learning (RL) inside learned world models (WMs) can post-train vision-language-action (VLA) policies with less physical interaction. However, WM-only training can exploit visual and dynamics prediction errors, degrading transfer to simulators and physical robots. Hybrid training with environment rollouts curbs this exploitation, yet still requires deciding how strongly WM rollouts should influence policy updates and how much environment interaction to retain. We introduce \ours, a \textbf{Re}play-\textbf{Cal}ibrated hybrid group-relative policy optimization (GRPO) method for \textbf{VLA} post-training that answers both questions. Under GRPO, a reward offset shared within a group cancels, but discrepancies in relative outcomes still distort the advantages used for policy updates. We therefore normalize advantages separately for environment and WM groups and scale WM advantages by a WM advantage weight, which sets the contribution of each WM rollout independently of the sampling proportion. At the end of each training stage, \ours{} replays recorded environment actions through the WM at no interaction cost. Regressing environment advantages on WM advantages sets the WM advantage weight, and the reverse regression, which measures how much environment signal the WM recovers, sets the next stage's environment fraction. A per-task allocation then directs environment rollouts toward tasks whose groups still have mixed outcomes. On four LIBERO suites, \ours improves mean success by 7.6 points over WM-only training and uses 43\% fewer simulator trajectories than simulator-only training at comparable success. On real-world cloth folding, it matches a fixed mixture with 45\% fewer robot episodes and comes within 2 points of robot-only training with 72\% fewer.
\par\smallskip\noindent\textup{Project page: \href{https://zhachang2.github.io/ReCal-VLA-page/}{zhachang2.github.io/ReCal-VLA-page}}
\end{abstract}

%% file: sec/1_intro.tex
\section{Introduction}
\label{sec:intro}

Vision-language-action (VLA) models learn versatile manipulation skills from large-scale robot demonstrations~\cite{brohan2023rt1,zitkovich2023rt2,oneill2024openx,octo2024,kim2024openvla,black2024pi0,intelligence2025pi05} and adapt to new tasks through supervised fine-tuning~\cite{chi2023diffusionpolicy,kim2025oft}. Reinforcement learning (RL), which has substantially improved large language models~\cite{ouyang2022instructgpt,shao2024deepseekmath,guo2025deepseekr1}, can further improve them from task feedback~\cite{lu2025vlarl,li2025simplevlarl,tan2025riptvla}, including experience collected during real-world deployment~\cite{luo2024hilserl,physicalintelligence2025recap}. The bottleneck is the experience itself. Physical rollouts consume robot time, resets and human supervision, while simulators require dedicated scene construction and leave a gap to the real world~\cite{tobin2017domain}. Reducing the interaction that RL needs is therefore central to scaling it on robots.

Learned world models (WMs) have long supplied imagined experience for model-based RL~\cite{sutton1991dyna,chua2018pets,hafner2020dreamer,hafner2023dreamerv3,wu2022daydreamer}, and video generative models can now simulate interactive visual environments~\cite{du2023unipi,yang2024unisim,bruce2024genie,agarwal2025cosmos}. An action-conditioned video WM can therefore generate rollouts for VLA post-training without executing them on a robot, and recent work shows that RL inside such models improves VLA policies~\cite{zhu2025wmpo,jiang2025world4rl,sharma2026worldgymnast}. The open question is how to use this experience when the model is imperfect.

Training on WM rollouts alone is unreliable. Inaccurate action responses and compounding prediction errors let a policy improve inside the WM without corresponding gains in the target environment (Fig.~\ref{fig:teaser}a)~\cite{janner2019mbpo,yu2020mopo,kidambi2020morel,jiang2026wovr}, much as a policy over-optimizes a learned reward model~\cite{skalse2022rewardhacking,gao2023overoptimization}. Existing approaches reduce the impact of WM errors through branched rollouts~\cite{janner2019mbpo,zhang2026vlambpo}, keyframe initialization~\cite{jiang2026wovr} and iterative policy--model improvement~\cite{guo2026vlaw}. A complementary remedy is to keep some interaction with the target environment, a simulator or the real robot; we call these \emph{environment rollouts}. Even a fraction of them anchors the update to true outcomes. Mixing the two sources, however, raises two questions that a fixed mixture leaves open: (i)~how much a WM rollout should count, and (ii)~how many environment rollouts each task still needs as the policy and the WM change during training.

We answer the first question by examining what group-relative policy optimization (GRPO)~\cite{shao2024deepseekmath} consumes. GRPO and related critic-free VLA post-training~\cite{tan2025riptvla} normalize outcomes within a group of rollouts from the same initial state. A reward offset shared by the whole group therefore cancels, whereas discrepancies in relative outcomes, such as a reordering that flips advantage signs (Fig.~\ref{fig:teaser}b) or a mixed group rendered uniform, still distort the advantages used for policy updates. This motivates calibrating WM contributions directly in advantage space. Outcome accuracy pooled across states mixes the two, and therefore gives an incomplete account of WM usefulness~\cite{kang2026drill}. Our hybrid update normalizes environment and WM groups separately, so that source-level offsets cancel, and scales WM advantages by a WM advantage weight calibrated in advantage space. The weight sets how much each WM rollout counts, not how many are sampled.

The second question depends on how reliable the WM currently is, which changes as the policy drifts away from the WM's training data~\cite{liu2026worldvlaloop}. We therefore introduce \ours, \textbf{Re}play-\textbf{Cal}ibrated hybrid GRPO for \textbf{VLA} post-training (Fig.~\ref{fig:method}). At the end of each training stage, \ours replays held-out environment actions through the WM without extra interaction~\cite{duan2026worldoracle}. The replay distinguishes spurious advantage variation introduced by the WM, such as mixed WM outcomes on a group whose true outcomes are uniform, from genuine variation that the WM fails to reproduce, such as a mixed true group that the WM renders uniform. The former calls for reducing the WM contribution to policy updates, while the latter calls for retaining more environment interaction. Two least-squares slopes between WM and true advantages, rather than a single reliability score, capture them separately: the forward slope $\lambda$ sets the WM advantage weight, and the reverse slope $\kappa$ sets the environment fraction of the next stage. A per-task allocation then directs environment rollouts toward tasks whose groups still have mixed outcomes. Unlike earlier adaptive schemes~\cite{lai2021autombpo,frauenknecht2024macura} and concurrent work that calibrates trust in imagined data under a fixed real-data budget~\cite{li2026merl}, \ours also decides how much environment data to collect.

On four LIBERO suites~\cite{liu2023libero}, \ours improves mean success by 7.6 points over WM-only training and uses 43\% fewer simulator trajectories than simulator-only training at comparable success (Fig.~\ref{fig:teaser}c). On real-world cloth folding, it reaches 84\% success, matching a fixed mixture with 45\% fewer robot episodes and coming within 2 points of robot-only training with 72\% fewer. Our contributions are threefold:
\begin{itemize}
    \item An analysis of what GRPO needs from a WM, where shared offsets cancel but discrepancies in relative outcomes distort the advantages, leading to a hybrid update with per-source normalization and a least-squares WM advantage weight.
    \item A replay-based calibration that estimates two regression slopes between WM and true advantages at no extra interaction cost, which set the WM advantage weight and the environment fraction, together with a per-task allocation of environment rollouts.
    \item Experiments in simulation and on a robot showing that calibrated WM rollouts can replace a large share of environment interaction. \ours improves over WM-only training on LIBERO and reaches success comparable to simulator-only and robot-only training with 43\% fewer simulator trajectories and 72\% fewer robot episodes.
\end{itemize}

%% file: sec/2_related.tex
\section{Related Work}
\label{sec:related}

\paragraph{RL for VLA policies.}
Critic-free RL with group-relative advantages has become a standard recipe for VLA post-training in simulation~\cite{lu2025vlarl,li2025simplevlarl,tan2025riptvla,rlinf,zang2025rlinfvla}, extends to flow-based VLAs~\cite{chen2025pirl}, and generalizes better than supervised fine-tuning alone~\cite{liu2025rlvlagen}. On real robots, actor-critic and advantage-conditioned methods improve policies from deployment experience~\cite{luo2024hilserl,physicalintelligence2025recap,li2025grrl}. These methods draw every rollout from the target environment. We keep the GRPO update and reduce the environment rollouts it needs.

\paragraph{RL in learned world models.}
Action-conditioned video models~\cite{du2023unipi,yang2024unisim,bruce2024genie,agarwal2025cosmos,guo2026ctrlworld,davies2026primesim} can act as learned simulators for robot policies. Some methods use them to score or rank actions~\cite{hung2025nora15}, while others run closed-loop VLA RL inside them~\cite{xiao2025worldenv,li2025vlarft,zhang2025prophrl,zhu2025wmpo,jiang2025world4rl,sharma2026worldgymnast,yang2026rise,wang2026rawdream}. As in model-based RL, a policy optimized purely in a WM can exploit its errors~\cite{janner2019mbpo,jiang2026wovr}. Other work therefore makes the WM more reliable, through keyframe initialization and online refinement with policy rollouts~\cite{jiang2026wovr}, or through iterative policy--model improvement~\cite{guo2026vlaw,liu2026worldvlaloop}. We also refresh the WM periodically, but our focus is how much its rollouts should count and how many environment rollouts are still needed.

\paragraph{Mixing real and imagined experience.}
Dyna-style methods train on imagined experience generated by a learned model~\cite{sutton1991dyna,chua2018pets,hafner2020dreamer,hafner2023dreamerv3}, and offline variants penalize model uncertainty~\cite{yu2020mopo,kidambi2020morel}. MBPO~\cite{janner2019mbpo} and VLA-MBPO~\cite{zhang2026vlambpo} branch short model rollouts from real states, while AutoMBPO~\cite{lai2021autombpo}, MACURA~\cite{frauenknecht2024macura} and learned reweighting of imagined transitions~\cite{huang2021reweight} adapt how model data is used. Closest to ours, MERL~\cite{li2026merl} weights the imagined data of a VLA by a trust estimate computed from visual and progress prediction errors on replayed actions, and updates the WM online under a fixed real-data budget. Paired real--WM comparisons~\cite{duan2026worldoracle} and policy-aligned WM training~\cite{kang2026drill} ask whether a WM preserves action outcomes. We measure agreement in advantage space, where shared offsets cancel, and use it to set the WM advantage weight and environment fraction.

\paragraph{Adaptive sampling.}
Dynamic sampling filters groups with uniform outcomes so that updates concentrate on informative prompts or states~\cite{yu2025dapo,li2025simplevlarl,tan2025riptvla}. Our per-task allocation applies this idea to environment rollouts: it keeps initial states uniform over tasks and changes only which source rolls out the secondary group.

%% file: sec/3_consistency.tex
\section{Method}
\label{sec:method}

\subsection{Preliminaries}
\label{sec:preliminaries}

\paragraph{GRPO for VLA.}
\label{subsec:grpo_vla}
We consider a VLA policy $\pi_\theta(a_h \mid o_h, l)$ initialized via supervised fine-tuning~\cite{kim2024openvla,kim2025oft}, which maps visual observations $o_h$ and a language instruction $l$ from a multi-task set $\mathcal{T}$ to action chunks $a_h$ over horizon $H$. Following critic-free VLA post-training~\cite{lu2025vlarl,li2025simplevlarl,tan2025riptvla} and distributed rollout systems such as RLinf~\cite{rlinf}, each sampled initial state $(o_0, l)$ spawns a group of $M$ trajectories $\{\tau_m\}_{m=1}^{M}$ with binary task-completion outcomes $\{y_m\}_{m=1}^{M} \in \{0, 1\}^M$. Group-relative policy optimization (GRPO)~\cite{shao2024deepseekmath} estimates each trajectory's advantage relative to its group peers:
\begin{equation}
    A_m = \frac{y_m - \mu_y}{\sigma_y + \epsilon},
    \label{eq:grpo_norm}
\end{equation}
where $\mu_y$ and $\sigma_y$ denote the within-group mean and standard deviation, and $\epsilon > 0$ ensures numerical stability. Two properties of Eq.~\eqref{eq:grpo_norm} anchor our analysis: any constant reward offset shared across a group cancels in $A_m$, and any uniform group where all $M$ rollouts succeed or fail ($\sigma_y = 0$) produces zero advantage and zero policy gradient.

Each trajectory contributes a clipped surrogate loss~\cite{schulman2017ppo,shao2024deepseekmath} averaged over action chunks:
\begin{equation}
\begin{aligned}
    \ell(\theta;\, A) &= -\frac{1}{H}\sum_{h=1}^{H} \min\!\Big(r_h(\theta) A, \\
    &\qquad \mathrm{clip}\big(r_h(\theta),\, 1 - \epsilon_{\mathrm{clip}},\, 1 + \epsilon_{\mathrm{clip}}\big) A\Big),
\end{aligned}
\label{eq:grpo_loss}
\end{equation}
where $r_h(\theta) = \pi_\theta(a_h \mid o_h, l) / \pi_{\theta_{\mathrm{old}}}(a_h \mid o_h, l)$ denotes the likelihood ratio between the current and rollout policies.

\paragraph{World-model rollouts.}
\label{subsec:wm_rm}
To reduce physical or simulator interaction, an action-conditioned video world model $f_\phi$ can serve as a learned visual simulator~\cite{zhu2025wmpo,jiang2025world4rl,sharma2026worldgymnast}. Starting from a real initial observation $o_0$ and instruction $l$, a diffusion video model $f_\phi$~\cite{peebles2023dit,wan2025} autoregressively predicts future visual observations $\hat{o}_{1:H}$ conditioned on policy action chunks $a_{0:H-1}$. Since generated videos carry no termination signal, a frozen visual reward model $R_\psi$ scores each generated trajectory to predict a binary outcome $\hat{y} \in \{0, 1\}$~\cite{jiang2026wovr,guo2026vlaw}. Generating $M$ trajectories from $(o_0, l)$ inside $(f_\phi, R_\psi)$ thus forms a WM group without executing actions in the target environment.

\paragraph{Challenges.}
\label{subsec:wm_challenges}
Errors of $(f_\phi, R_\psi)$ cause both state-level reward offsets and within-group errors in relative outcomes~\cite{janner2019mbpo,jiang2026wovr,kang2026drill}; offsets cancel only within single-source groups, whereas relative-outcome errors distort the advantages. WM reliability also changes as $\pi_\theta$ drifts from the training data of $f_\phi$~\cite{liu2026worldvlaloop}, and environment rollouts on solved or intractable tasks give no gradient~\cite{yu2025dapo}.

%% file: sec/4_method.tex
\begin{figure*}[t]
  \centering
  \includegraphics[width=0.95\textwidth]{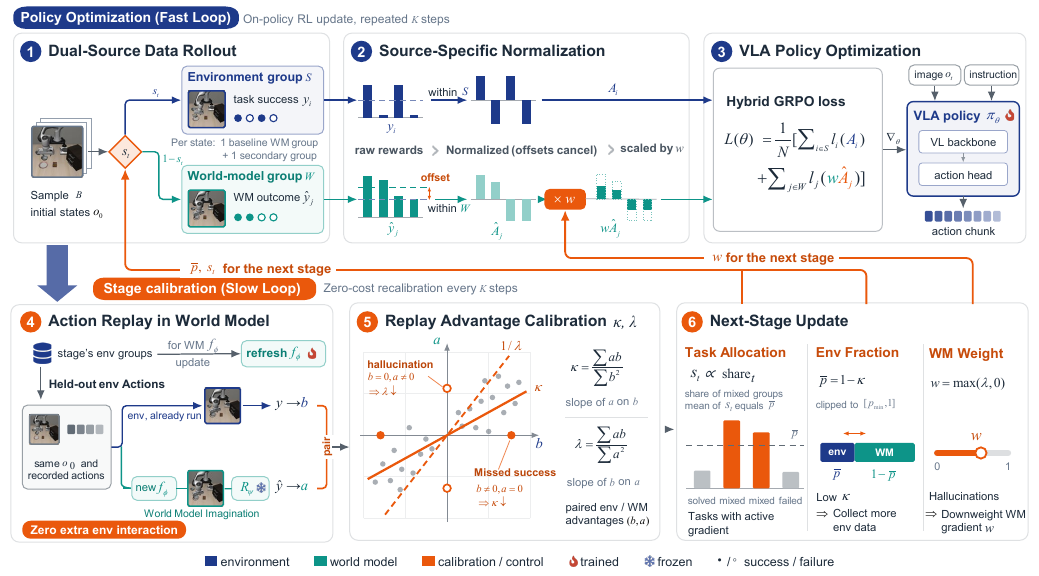}
  \caption{\textbf{Overview of \ours.} Top: one hybrid GRPO step (Sec.~\ref{subsec:hybrid_grpo}). Bottom: stage calibration (Secs.~\ref{subsec:replay_calibration}--\ref{subsec:task_allocation}). Held-out environment actions are replayed through the WM to estimate $\kappa$ and $\lambda$, which set $w$, $\bar p$ and $s_t$ for the next stage.}
  \label{fig:method}
\end{figure*}

We present \ours, replay-calibrated hybrid GRPO, which couples a hybrid policy update with periodic stage-end calibration. Within each stage, the policy optimizes a hybrid objective that normalizes advantages separately per source and scales WM advantages by a replay-calibrated weight; at fixed stage boundaries, we replay held-out environment actions through the world model, optionally refreshed, to update the WM advantage weight, the environment fraction, and the per-task environment probabilities. Fig.~\ref{fig:method} gives the overview.

\subsection{Hybrid GRPO with per-source normalization}
\label{subsec:hybrid_grpo}

At each optimization step, we collect a hybrid batch of environment trajectories $\mathcal{S}$ and WM trajectories $\mathcal{W}$, organized into groups of size $M$. To prevent source-level reward offsets from contaminating zero-variance groups or flipping advantage signs across sources, we construct groups strictly within each source and normalize environment and WM groups separately.

While separate normalization removes additive offsets between sources, imperfect dynamics inside $f_\phi$ still introduce noise in relative outcomes within WM groups. We therefore scale each WM trajectory's group-normalized advantage $A_j$ by a non-negative WM advantage weight $w \ge 0$ before averaging the clipped losses over all $N = |\mathcal{S}| + |\mathcal{W}|$ rollouts:
\begin{equation}
    \mathcal{L}(\theta) = \frac{1}{N}\left[\sum_{i \in \mathcal{S}} \ell_i(\theta;\, A_i) + \sum_{j \in \mathcal{W}} \ell_j(\theta;\, w A_j)\right].
    \label{eq:hybrid_loss}
\end{equation}
Because $w \ge 0$ scales each WM advantage linearly inside the clipped surrogate objective, it directly sets the gradient magnitude of individual WM trajectories relative to environment trajectories, decoupling the per-trajectory weight of WM rollouts from the sampling ratio $|\mathcal{S}| : |\mathcal{W}|$.

\subsection{Replay-based advantage calibration}
\label{subsec:replay_calibration}

To track evolving policy distributions, the WM can be refreshed at the end of each fixed-length training stage. In that case, we partition the environment groups collected during the stage into a model-training split and a disjoint held-out calibration split, both already used for policy updates, and fine-tune the video world model $f_\phi$ on the model-training split mixed with historical environment trajectories to mitigate forgetting~\cite{lin1992replay,guo2026vlaw}; without a refresh, all environment groups are held out. We then replay the initial observations and exact action sequences from the held-out split through $f_\phi$ and the frozen reward model $R_\psi$, obtaining paired true and predicted outcomes $(y, \hat{y})$ with zero additional environment interaction~\cite{duan2026worldoracle}.

Within each replayed group, we normalize $y$ and $\hat{y}$ separately to obtain paired true advantages $b$ and WM advantages $a$, removing state-level difficulty offsets that cancel under GRPO. Since $a$ and $b$ have zero mean within each group, we compute two least-squares regression slopes through the origin over all replayed trajectories:
\begin{equation}
    \kappa = \frac{\sum a\,b}{\sum b^2}, \quad \lambda = \frac{\sum a\,b}{\sum a^2}, \quad r^2 = \kappa\lambda = \frac{\left(\sum a\,b\right)^2}{\sum a^2\,\sum b^2}.
    \label{eq:slopes}
\end{equation}
Here $\lambda$ is the forward regression slope of $b$ on $a$, measuring the expected true advantage per unit of WM advantage. Conversely, $\kappa$ is the reverse regression slope of $a$ on $b$, measuring the WM advantage reproduced per unit of true advantage, while their product $r^2 = \kappa\lambda$ equals the fraction of true advantage variance explained by the WM advantages.

We set the WM advantage weight $w$ in Eq.~\eqref{eq:hybrid_loss} to minimize the relative mean-squared error between scaled WM advantages $w a$ and true advantages $b$:
\begin{equation}
\begin{aligned}
    E(w) &= \frac{\sum (w\,a - b)^2}{\sum b^2} = w^2\,\frac{\kappa}{\lambda} - 2w\kappa + 1, \\
    w^* &= \arg\min_w E(w) = \lambda, \quad E(\lambda) = 1 - \kappa\lambda.
\end{aligned}
\label{eq:mse_weight}
\end{equation}
For $\sum a^2 > 0$, using unscaled WM advantages ($w = 1$) yields $E(1) > 1$ whenever $\sum a^2 > 2\sum ab$ (equivalently, $\lambda < 1/2$), a larger advantage error than discarding WM advantages altogether ($E(0) = 1$). Setting $w = \max(\lambda, 0)$ guarantees $E(w) \le 1$ on the replayed pairs. This bounds the error of the scalar advantages rather than of the policy gradient, which also depends on the score vectors $\nabla_\theta \log \pi_\theta$; moreover, the replay is open-loop on recorded environment actions, whereas WM groups are generated closed-loop during training. We therefore use $w$ as a calibrated rescaling of WM advantages rather than an exact correction.

\subsection{Environment fraction and per-task allocation}
\label{subsec:task_allocation}

While $\lambda$ down-weights spurious WM variance in the loss, setting the next stage's environment fraction from the residual $1 - \kappa\lambda$ would penalize the same variance a second time. Indeed, spurious WM variance on states with uniform true outcomes ($b = 0$) lowers $\lambda$ without lowering $\kappa$, whereas missed WM signal on states with mixed outcomes ($b \neq 0, a = 0$) lowers $\kappa$ even when $\lambda$ remains high. We therefore set data collection from $\kappa$~\cite{lai2021autombpo,frauenknecht2024macura}. Each step samples $B$ initial states, giving $G = 2B$ groups: every initial state has a primary WM group and a secondary group that runs either in the environment or in the WM. We set the environment fraction $\bar{p}$, the mean probability that a secondary group runs in the environment, and the expected number of environment groups per step $n_{\mathrm{env}}$ as:
\begin{equation}
    \bar{p} = \min\!\big(1,\; \max(p_{\min},\, 1 - \kappa)\big), \qquad n_{\mathrm{env}} = B\,\bar{p},
    \label{eq:env_fraction}
\end{equation}
a heuristic schedule that moves from a 1:1 split of groups when the WM reproduces no true advantage ($\kappa \le 0$) down to a floor $p_{\min} > 0$ that sustains continual WM updates and replay calibration as $\kappa \to 1$. Prior to the first replay, training initializes with $\bar{p} = 1$ and $w = 1$.

Finally, environment groups on solved or intractable tasks are wasted, since their uniform outcomes ($\sigma_y = 0$) give no gradient. To avoid them without changing the task distribution, we sample initial states uniformly across all $|\mathcal{T}|$ tasks and adapt only the per-task environment probabilities $s_t \in [0, 1]$. Of the $n_t$ environment groups executed on task $t \in \mathcal{T}$ in the preceding stage, let $i_t$ count the mixed groups ($\sigma_y > 0$). Using Laplace smoothing, we solve for $s_t$ under the environment fraction $\bar{p}$:
\begin{equation}
\begin{gathered}
    \mathrm{share}_t = \frac{i_t + 1}{n_t + 2}, \quad s_t = \min(1,\, c\cdot\mathrm{share}_t), \\
    \frac{1}{|\mathcal{T}|}\sum_{t \in \mathcal{T}} s_t = \bar{p},
\end{gathered}
\label{eq:task_allocation}
\end{equation}
where $c \ge 0$ is the smallest value satisfying the constraint, found by bisection. For an initial state of task $t$, the secondary group runs in the environment with probability $s_t$ and in the WM otherwise, concentrating environment interaction on tasks with mixed outcomes while solved tasks stay covered by WM groups.

%% file: sec/5_experiments.tex
\section{Experiments}
\label{sec:exp}

We test whether replay calibration saves environment interaction without losing task success. On LIBERO, we compare against WM-only training (Sec.~\ref{subsec:main_results}), measure the saving against simulator-only training and a fixed mixture, and track how $(\kappa,\lambda)$ adapt (Sec.~\ref{subsec:calibration_analysis}). We then ablate the components (Sec.~\ref{subsec:ablations}) and test perturbed scenes and a physical robot (Secs.~\ref{subsec:libero_pro}--\ref{subsec:real_world}).

\subsection{Experimental setup}
\label{subsec:exp_setup}

\begin{figure}[t]
\centering
\includegraphics[height=2.45cm]{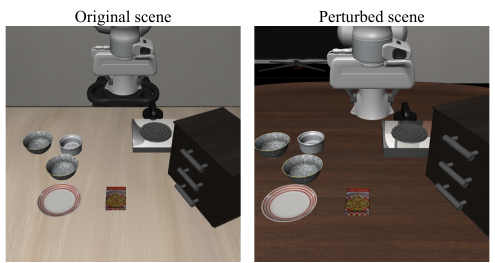}\hfill
\includegraphics[height=2.45cm]{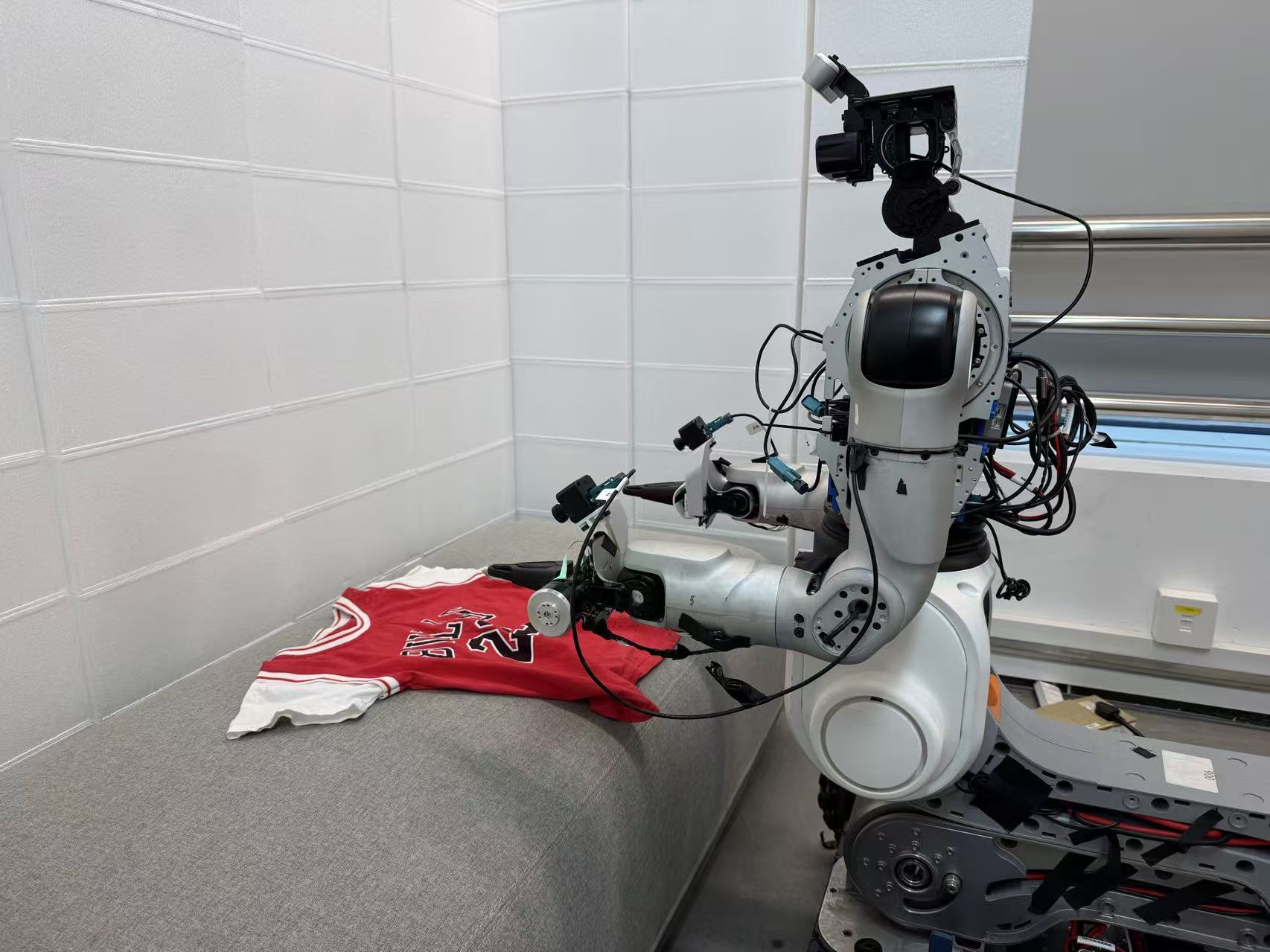}
\caption{\textbf{Evaluation scenes beyond LIBERO.} (a)~LIBERO-PRO: original (left) and perturbed (right) scene. (b)~Real-world setup with three cameras.}
\label{fig:scenes}
\end{figure}

\paragraph{Benchmark and models.}
We use the four LIBERO suites~\cite{liu2023libero}, each with ten tasks, and the official evaluation protocol. Following the low-data setting of WMPO~\cite{zhu2025wmpo}, every method starts from the same one-demonstration OpenVLA-OFT checkpoint~\cite{kim2025oft}. The WM and reward model are the released suite-specific Wan2.2-TI2V-5B~\cite{wan2025} action-conditioned models and frozen visual success classifiers in RLinf~\cite{rlinf}.

\paragraph{Baselines and protocol.}
All methods train for 300 steps with 256 trajectories per step in groups of eight. \emph{WM-Only} draws all trajectories from the WM. \emph{Fixed Hybrid} splits every step equally between 128 simulator and 128 WM trajectories. \emph{Sim-Only} uses simulator trajectories alone, either 128 per step, matching the initial simulator share of the hybrid methods, or 256. \ours refreshes the WM and recalibrates every 20 steps; its remaining hyperparameters are in the supplementary material.

\paragraph{Metrics.}
We count the simulator trajectories used for training and exclude evaluation. We report success at a matched simulator budget, the simulator trajectories needed to first reach a target success rate, and the final success together with the trajectories it consumed and the success gain per 10k simulator trajectories.

\begin{table}[t]
\centering
\footnotesize
\setlength{\tabcolsep}{4.8pt}
\input{gen/tab_main}
\caption{\textbf{Main comparison on LIBERO} (success rate, \%). All methods start from the same one-demonstration OpenVLA-OFT checkpoint and use the standard LIBERO evaluation protocol. The OpenVLA-OFT, Online GRPO and WMPO rows are the results reported by WoVR~\cite{jiang2026wovr}.}
\label{tab:main}
\end{table}

\begin{figure*}[t]
\centering
\includegraphics[width=0.9\textwidth]{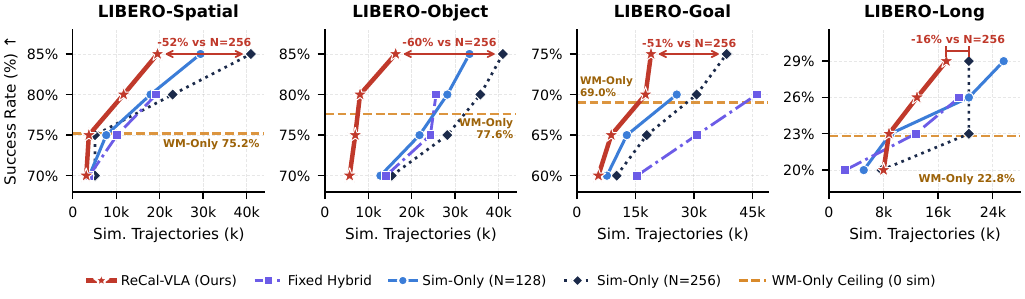}
\caption{\textbf{Simulator trajectories needed to reach each success milestone} (upper-left is better; Fixed Hybrid on Goal is extended beyond 300 steps to measure the budget required to reach 70\%). Dashed amber lines: \emph{WM-Only} after 300 steps.}
\label{fig:frontier}
\end{figure*}

\subsection{Main results on LIBERO}
\label{subsec:main_results}

\paragraph{Hybrid rollouts break the pure-WM ceiling.}
Table~\ref{tab:main} compares \ours against \emph{WM-Only} under the same protocol, with reported results of earlier methods for reference. Training purely inside our video world model (\emph{WM-Only}) reaches a $61.2\%$ four-suite mean, yet hits a clear transfer ceiling on every suite. Anchoring WM rollouts with replay-calibrated environment interaction, \ours lifts success on all four suites and reaches a $68.8\%$ mean, $7.6$ points above \emph{WM-Only}.

\begin{table*}[t]
\centering
\footnotesize
\setlength{\tabcolsep}{2.45pt}
\input{gen/tab_efficiency}
\caption{\textbf{Sample efficiency against simulator-only RL.} Left: success rate at a matched same simulator budget. Middle: simulator trajectories to reach the target success rate. Right: 300-step success, simulator trajectories, savings relative to \emph{Sim-Only} ($N{=}128$), and gain over the SFT policy ($40.5\%$) per $10\mathrm{k}$ simulator trajectories. $^{*}$\,Reaches $73.0\%$ after $38.4\mathrm{k}$ trajectories. $^{\dagger}$\,Uses $3.5\times$ the simulator trajectories of \ours.}
\label{tab:efficiency}
\end{table*}

\subsection{Sample efficiency and calibration dynamics}
\label{subsec:calibration_analysis}

\paragraph{Sample efficiency over \emph{Sim-Only} and \emph{Fixed Hybrid}.}
Table~\ref{tab:efficiency} and Fig.~\ref{fig:frontier} evaluate \ours on sample-efficiency metrics against pure simulator RL and fixed mixing. First, at a \textbf{matched $\sim\!20.5\mathrm{k}$ simulator budget}, \ours reaches a $68.8\%$ mean, against $60.8\%$ for \emph{Sim-Only} ($N{=}128$) and $59.4\%$ for \emph{Sim-Only} ($N{=}256$), confirming that calibrated WM rollouts, rather than a larger batch, drive the gain. Second, across \textbf{progressive target milestones} (Fig.~\ref{fig:frontier}), \emph{Sim-Only} ($N{=}256$) is the most data-hungry baseline, whereas \ours occupies the upper-left frontier on every suite: relative to \emph{Sim-Only} ($N{=}128$), \ours needs $71\%$ fewer simulator trajectories to reach $80\%$ on Object, and reaches $75\%$ on Goal, where \emph{Sim-Only} ($N{=}128$) stalls at $73.0\%$. At 300 steps, \ours surpasses \emph{Sim-Only} ($N{=}128$), $68.8\%$ vs.\ $67.7\%$, while saving $43.4\%$ of simulator trajectories and nearly doubling the success gain per $10\mathrm{k}$ simulator trajectories.

\begin{figure*}[t]
\centering
\includegraphics[width=0.9\textwidth]{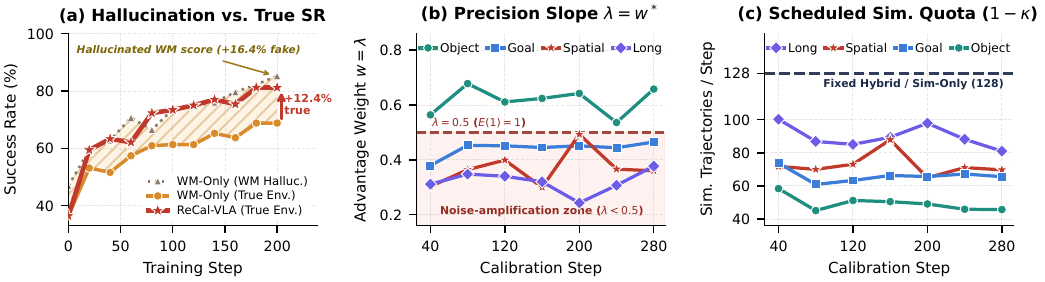}
\caption{\textbf{Calibration dynamics.} (a)~LIBERO-Object: WM-judged success of \emph{WM-Only} (dashed), its true success, and \ours. (b)~Replay-estimated $\lambda = w^*$; shaded: $\lambda < 0.5$, where $w = 1$ gives $E(1) > 1$ (Eq.~\eqref{eq:mse_weight}). (c)~Simulator trajectories per step set by $1 - \kappa$, relative to the 128 of \emph{Fixed Hybrid}.}
\label{fig:calibration}
\end{figure*}

\paragraph{Suppressing WM hallucination and adapting the environment fraction across suites.}
Fig.~\ref{fig:calibration} explains why \ours achieves these gains without manual ratio tuning. First, on LIBERO-Object (Fig.~\ref{fig:calibration}a), optimizing purely inside $(f_\phi, R_\psi)$ inflates WM-judged success to around $85\%$ while true success stalls at $68.8\%$; 
anchoring updates with calibrated environment rollouts in \ours lifts true task success by $+12.4$ points over \emph{WM-Only} by step 200. Second, the replay slope $\kappa$ sets the per-step simulator count of each suite according to how well its WM reproduces true advantages (Fig.~\ref{fig:calibration}c).
On Object, where $\kappa$ is high, \ours lowers simulator collection to $45\text{--}58$ trajectories/step and reaches $80\%$ with $68\%$ fewer simulator trajectories than \emph{Fixed Hybrid}.
On Spatial and Goal, a moderate $\kappa$ keeps collection mostly at $61\text{--}74$ trajectories/step (briefly peaking at 89 on Spatial at step 160), $30\text{--}52\%$ below \emph{Fixed Hybrid}. 
On 512-step Long, where $\kappa$ drops for long-horizon videos, \ours automatically retains $81\text{--}100$ simulator trajectories/step to sustain policy progress.

\begin{table}[t]
\centering
\footnotesize
\setlength{\tabcolsep}{3.0pt}
\input{gen/tab_ablation}
\caption{\textbf{Component ablation on LIBERO-Spatial} (300 steps). Starting from \emph{WM-Only}, we add environment rollouts with the $\kappa$-based environment fraction, the $\lambda$-based WM advantage weight, and per-task allocation.}
\label{tab:ablation}
\end{table}

\subsection{Ablation study and bottleneck tasks}
\label{subsec:ablations}

\paragraph{Controlled component-by-component ablation.}
Table~\ref{tab:ablation} isolates each core mechanism of \ours on LIBERO-Spatial at 300 steps. Starting from \emph{WM-Only} at $75.2\%$, adding environment rollouts with the $\kappa$-based environment fraction lifts task success by $5.4$ points to $80.6\%$. However, the replay slope on Spatial stays at $\lambda < 0.5$ throughout training (Fig.~\ref{fig:calibration}b), where unscaled WM advantages have a larger advantage error than dropping them, $E(1) > 1$ in Eq.~\eqref{eq:mse_weight}. Setting $w = \max(\lambda, 0)$ further reduces simulator consumption from $22.9\mathrm{k}$ to $21.5\mathrm{k}$ while slightly improving success to $81.4\%$. Finally, at the same simulator budget, per-task allocation adds $3.6$ points, bringing full \ours to $85.0\%$.

\begin{figure}[t]
\centering
\includegraphics[width=0.7\columnwidth]{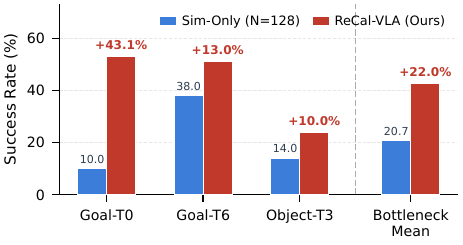}
\caption{\textbf{Breakthroughs over \emph{Sim-Only} ($N{=}128$) on stagnant bottleneck tasks} (300 steps). Bars show all tasks across Spatial, Object, and Goal where \emph{Sim-Only} ($N{=}128$) fails to reach $50\%$ success after $38.4\mathrm{k}$ simulator trajectories, alongside their mean.}
\label{fig:pertask}
\end{figure}

\paragraph{Unlocking stagnant bottleneck tasks.}
Fig.~\ref{fig:pertask} shows that \ours breaks the stagnation of simulator-only training on bottleneck tasks. On hard tasks of different environments, \emph{Sim-Only} ($N{=}128$) stays below $50\%$ even after $38.4\mathrm{k}$ simulator trajectories. This is consistent with per-task allocation: once other tasks saturate, Eq.~\eqref{eq:task_allocation} moves simulator groups to these tasks, whose groups still have mixed outcomes, while solved tasks stay in training through WM groups. Mean success on the three tasks more than doubles, from $20.7\%$ to $42.7\%$, with $47.7\%$ fewer simulator data.

%% file: gen/tab_main.tex
\begin{tabular}{@{}lccccc@{}}
\toprule
Method & Spatial & Object & Goal & Long & Avg. \\
\midrule
OpenVLA-OFT~\cite{kim2025oft} & 63.6 & 36.4 & 48.2 & 13.8 & 40.5 \\
+ Online GRPO~\cite{shao2024deepseekmath} & 66.6 & 45.2 & 52.2 & 14.6 & 44.6 \\
+ WMPO~\cite{zhu2025wmpo} & 67.8 & 65.4 & 56.6 & 13.8 & 50.9 \\
+ WM-Only & \underline{75.2} & \underline{77.6} & \underline{69.0} & \underline{22.8} & \underline{61.2} \\
\textbf{+ \ours (Ours)} & \textbf{85.0} & \textbf{84.6} & \textbf{77.4} & \textbf{28.2} & \textbf{68.8} \\
\bottomrule
\end{tabular}

%% file: gen/tab_efficiency.tex
\begin{tabular}{@{}lcccccccccccc@{}}
\toprule
& \multicolumn{4}{c}{SR @ Matched Budget (\%) $\uparrow$} & \multicolumn{4}{c}{Sim.\ Trajectories to Target SR ($\mathrm{k}$) $\downarrow$} & \multicolumn{4}{c}{300-Step Efficiency (4-Suite Mean)} \\
\cmidrule(lr){2-5}\cmidrule(lr){6-9}\cmidrule(lr){10-13}
Method & Spat. & Obj. & Goal & Long & Spatial \,$85\%$ & Object \,$80\%$ & Goal\,$75\%$ & Long\,$29\%$ & Ref.\ SR & Sim.\ $\downarrow$ & Saved $\uparrow$ & $\Delta\text{SR}/10\mathrm{k}\uparrow$ \\
\midrule
Sim-Only ($N{=}256$) & 73.8 & 71.9 & \underline{66.4} & \underline{25.4} & 41.0k & 35.8k & \underline{38.4k} & \underline{20.5k} & 77.2$^{\dagger}$ & 76.8k & $-100\%$ & $+4.8$ \\
Sim-Only ($N{=}128$) & \underline{79.3} & \underline{73.0} & \underline{66.4} & 24.5 & \underline{29.4k} & \underline{28.2k} & Fail$^{*}$ & 25.6k & 67.7 & \underline{38.4k} & $0.0\%$ & \underline{$+7.1$} \\
\textbf{\textsc{ReCal-VLA} (Ours)} & \textbf{85.0} & \textbf{84.6} & \textbf{77.4} & \textbf{28.2} & \textbf{19.5k} & \textbf{8.1k} & \textbf{19.0k} & \textbf{17.2k} & 68.8 & \textbf{21.8k} & \textbf{43.4\%} & $\mathbf{+13.0}$ \\
\bottomrule
\end{tabular}

%% file: gen/tab_ablation.tex
\begin{tabular}{@{}lcccccc@{}}
\toprule
Variant & Env. & $\kappa$ & $\lambda$ & Alloc. & SR (\%) & Sim. \\
\midrule
WM-Only & -- & -- & -- & -- & 75.2 & 0.0k \\
+ Env.\ \& $\kappa$ ($w{=}1$) & \checkmark & \checkmark & -- & -- & 80.6 & 22.9k \\
+ $\lambda$ weight ($w{=}\max(\lambda,0)$) & \checkmark & \checkmark & \checkmark & -- & \underline{81.4} & \textbf{21.5k} \\
\textbf{+ Per-task allocation (ours)} & \checkmark & \checkmark & \checkmark & \checkmark & \textbf{85.0} & \textbf{21.5k} \\
\bottomrule
\end{tabular}

%% file: sec/5b_pro_real.tex
\providecommand{\figph}[2]{\fbox{\parbox[c][#2][c]{0.96\linewidth}{\centering\small\textcolor{gray}{#1}}}}

\begin{figure*}[t]
\centering
\IfFileExists{fig/real_robot_wm.pdf}{\includegraphics[width=0.9\linewidth]{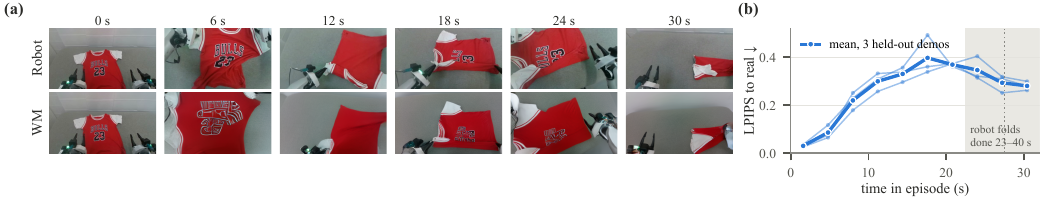}}{\includegraphics[width=0.9\linewidth]{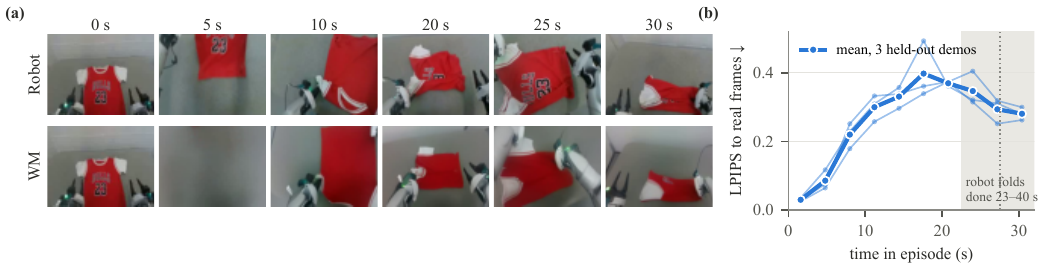}}
\caption{\textbf{The WM on cloth} (head camera). (a)~A successful robot episode (top) and the WM driven by its recorded actions from the same first frame (bottom). (b)~LPIPS to the real frames on three held-out demonstrations; shaded: robot folds finish (23--40\,s).}
\label{fig:real_cloth}
\end{figure*}

\subsection{Perturbed scenes: LIBERO-PRO}
\label{subsec:libero_pro}
LIBERO-PRO~\cite{zhou2025liberopro} moves every task into a perturbed scene (Fig.~\ref{fig:scenes}a). As in MERL~\cite{li2026merl}, we start from the OpenVLA-OFT~\cite{kim2025oft} initialization and evaluate on initial states not seen in training. We align the perturbation type, the four suites and the evaluation protocol with MERL and list its reported numbers for reference. Since LIBERO-PRO has no demonstrations, the WM and reward model are fine-tuned on rollouts of this policy in the perturbed scenes; training then follows Sec.~\ref{subsec:exp_setup}.

\begin{table}[t]
\centering\footnotesize\setlength{\tabcolsep}{5pt}
\begin{tabular}{@{}llc@{}}
\toprule
Method & Paradigm & Avg.\ SR (\%) $\uparrow$ \\
\midrule
SimpleVLA-RL~\cite{li2025simplevlarl} & Model-free        & 77.1 \\
WMPO$^\dagger$~\cite{zhu2025wmpo}      & Static model-based & 75.4 \\
WoVR$^\dagger$~\cite{jiang2026wovr}& Online-updated     & 77.8 \\
MERL~\cite{li2026merl}          & Trust-calibrated   & 79.7 \\
\midrule
\ours                           & Replay-calibrated  & 81.9 \\
\bottomrule
\end{tabular}
\caption{\textbf{LIBERO-PRO} (environment perturbation), mean success over the four suites. Baselines are taken from MERL; $^\dagger$: ported by MERL.}
\label{tab:pro}
\end{table}
\ours\ reaches 83.9, 87.5, 85.9 and 70.3\% on Spatial, Object, Goal and Long, an $81.9\%$ average, compared with the $79.7\%$ average reported by MERL (Tab.~\ref{tab:pro}).

\subsection{Real-world cloth folding}
\label{subsec:real_world}

\paragraph{Setup.}
A dual-arm robot folds a T-shirt placed by hand, and the reward is a binary outcome at the end of the episode (Fig.~\ref{fig:scenes}b). The policy is a flow-matching~\cite{lipman2023flow} VLA with a Qwen2.5-VL-3B~\cite{bai2025qwen25vl} backbone, fine-tuned on teleoperated demonstrations; its RL fine-tuning follows $\pi_{\texttt{RL}}$~\cite{chen2025pirl} in RLinf~\cite{rlinf}. The WM is PrimeSim~\cite{davies2026primesim}, which trains Wan2.2-TI2V-5B~\cite{wan2025} on large-scale cross-embodiment demonstrations; we further fine-tune it on our cloth-folding demonstrations. Fig.~\ref{fig:real_cloth} shows its drift. Each round collects 16 episodes in groups of four and performs one GRPO update, so a round is one stage of Sec.~\ref{sec:method} with two initial states. \emph{Real-Only}, \emph{WM-Only} and \emph{Fixed Hybrid} take the 16 episodes from the robot, the WM, or both equally. \ours\ starts from 8 robot and 8 WM episodes and recalibrates after every round. As replay showed the fixed WM's outcomes staying consistent with robot outcomes, the WM is never refreshed; all robot episodes are thus held out, and $\kappa$ and $\lambda$ are estimated on those accumulated so far. We evaluate the checkpoints after rounds 3, 6 and 10 with 50 robot episodes each.

\begin{table}[t]
\centering\footnotesize\setlength{\tabcolsep}{4pt}
\begin{tabular}{@{}lcc@{}}
\toprule
Setting & Robot episodes used in training & Success (\%) \\
\midrule
SFT            & --  & 58 \\
WM-Only        & --  & 68 \\
Real-Only      & 160 & \textbf{86} \\
Fixed Hybrid   & 80  & 84 \\
\ours          & \textbf{44} & \textbf{84} \\
\bottomrule
\end{tabular}
\caption{\textbf{Real-world cloth folding} after 10 rounds; success over 50 robot episodes, one run per setting.}
\label{tab:real}
\end{table}

\begin{figure}[t]
\centering
\IfFileExists{fig/real_success_vs_robot_eps.pdf}{\includegraphics[width=0.75\linewidth]{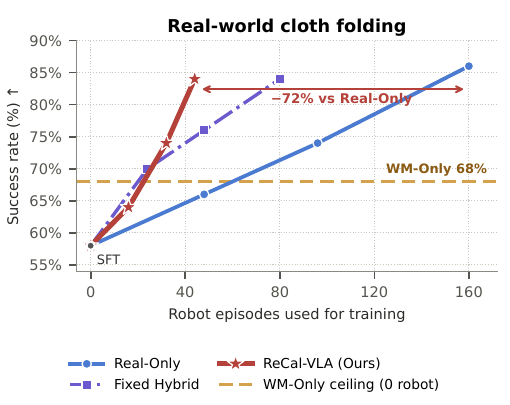}}{\figph{success vs.\ robot episodes: upload fig/real\_success\_vs\_robot\_eps.pdf}{4.0cm}}
\caption{\textbf{Success vs.\ robot episodes} (upper-left is better). Checkpoints after rounds 3, 6 and 10, 50 episodes each; dashed: WM-Only.}
\label{fig:real_curve}
\end{figure}

\paragraph{Results.}
Real-Only raises success from $58\%$ to $86\%$, and WM-Only reaches $68\%$ without any robot episode (Tab.~\ref{tab:real}). Fixed Hybrid and \ours\ both reach $84\%$, but \ours\ uses $45\%$ fewer robot episodes than Fixed Hybrid and $72\%$ fewer than Real-Only. Its robot share follows the calibrated environment fraction, using 4 or 8 robot episodes per round. At a similar robot budget (Fig.~\ref{fig:real_curve}), \ours\ reaches $84\%$, against $66\%$ for Real-Only and $76\%$ for Fixed Hybrid. A round of \ours\ takes about 40\,min vs.\ 70\,min for Real-Only, as WM episodes run in parallel.

%% file: sec/6_conclusion.tex
\section{Conclusion}
\label{sec:conclusion}

We studied hybrid RL for VLA post-training that mixes environment and world-model rollouts. Under GRPO, reward offsets shared within a group cancel, whereas discrepancies in relative outcomes distort the advantages. \ours therefore normalizes advantages per source and replays held-out environment actions to set the WM advantage weight and the environment fraction. It matches simulator-only training on LIBERO with $43\%$ fewer simulator trajectories, and comes within 2 points of robot-only training on real cloth folding with $72\%$ fewer robot episodes.

\section*{LLM Usage Statement}
Large language models (Claude, Codex) were used to help write and debug the experiment code and to draft and edit parts of the manuscript text from the authors' notes. All experimental design, results, analysis and claims were decided and checked by the authors, who take responsibility for the full content including citations.